\documentclass{article} 
\usepackage{nips15submit_e,times}
\usepackage{hyperref}
\usepackage{url}
\usepackage{authblk}
\usepackage{amsmath,graphicx,booktabs}
\usepackage{epstopdf,amssymb}
\usepackage{multirow}
\usepackage{fixltx2e}
\usepackage{subfigure}
\usepackage{float}
\usepackage{titlesec}

\title{\textbf{DASN:  Data-Aware Skilled Network for Accurate MR Brain Tissue Segmentation}} 
\author{Yang Deng$^{1,2}$, Yao Sun$^{1,2}$, Yongpei Zhu$^{1,2}$, Shuo Zhang$^{2}$, Mingwang Zhu$^{3}$,Kehong Yuan$^{1*}$\\
	$^{1}$Graduate School at Shenzhen, Tsinghua University, Shenzhen 518055, China.\\
	$^{2}$Department of Biomedical Engineering, Tsinghua University, Beijing 100084, China.\\
	$^{3}$Beijing Sanbo Brain Hospital, Beijing 100825, China.\\
	*Corresponding author:Kehong Yuan (e-mail:yuankh@sz.tsinghua.edu.cn)
}

\nipsfinalcopy 

\begin{document}

\maketitle

\begin{abstract}
Accurate segmentation of MR brain tissue is a crucial step for diagnosis, surgical planning, and treatment of brain abnormalities. Automatic and reliable segmenta-tion methods are required to assist doctor. Over the last few years, deep learning especially deep convolutional neural networks (CNNs) have emerged as one of the most prominent approaches for image recognition problems in various do-mains. But the improvement of deep networks always needs inspiration, which is rare for the ordinary. Until now, there have been reasonable MR brain tissue segmentation methods, all of which can achieve promising performance. These different methods have their own characteristic and are distinctive for data sets. In other words, different models performance vary widely on the same data sets and each model has what it is skilled in. It is on the basis of this, we propose a judgement to distinguish data sets that different models are good at. With our method, the segmentation accuracy can be improved easily based on the existing models, neither without increasing training data nor improving the network. We validate our method on the widely used IBSR 18 dataset and obtain average dice ratio of 88.06\%, while it is 85.82\% and 86.92\% when only using separate one model respectively.\\

\end{abstract}

\section{Introduction}
Both the central nervous system degenerative disease and epilepsy are related to the morphological changes in the brain tissue. The accurate segmentation of the brain tissue is the first step in the volume and quantitative analysis of the brain. It is im-portant for the diagnosis and treatment of brain diseases, especially the neurodegen-erative diseases.\\
 	
MRI can be more clearly and safer to display the structure of the brain because of its non-invasive, non-radioactive, free selection profile, higher signal to noise ratio, and higher resolution of the soft tissue with smaller density difference, so as to provide more information for the pathological diagnosis of brain diseases. It has become a common method for the examination of brain diseases. In medical institutions, when doctors get the MRI images of the patients, they often need to judge whether the patient's brain atrophies by their experience. When dividing the MRI brain tissue, it is necessary to sketch each slice, which is not only time-consuming, but also very tiring. At the same time, it is easy to produce fatigue error. On the other hand, in remote areas and hospitals, with insufficient medical resources and lack of experienced doc-tors, segmentation of MRI images has become a difficult problem.  \\	

Computer aided diagnostic technology (CAD) is used to help doctors diagnose pa-tients through a computer, thus reducing the burden for doctors and improving the efficiency and accuracy of diagnosis. This technique has been applied to a variety of medical diagnosis problems, and an automatic segmentation method which can pro-vide high precision near the expert segmentation standard, is urgently needed to assist the doctor.\\	

The traditional MR brain tissue segmentation algorithms\cite{Greenspan2006Constrained}\cite{Despotovi2015MRI}\cite{Qin2017Brain} \cite{Marroquin2002An}use artificial de-signed features and classifies the extracted features. However, the gray level changes of the MR brain images are large and the gray values of different types are not very different. This image feature leads to the complexity of features extraction and the difficulties to artificially design satisfactory features, which makes the segmentation results still have long way to go.\\

With the popularity of deep learning algorithms in the field of computer vision, more and more people begin to use convolution neural networks to classify, detect and segment images, which have achieved shocking effects in the field of natural images. The same work is being done in medical field and has become a hot topic.\\

Many scholars have proposed many MR image segmentation methods based on convolution neural network \cite{Zhang2015Deep}\cite{Nie2015FULLY}\cite{Moeskops2016Automatic} \cite{Bao2015Multi}and these methods have also achieved good results. But without exception, these methods performances are very different in different data sets. In other words, different models performance vary widely on the same data sets and each model has what it is skilled in. It is of great significance in medical field of data scarcity that how to exploit the aspects that different methods are good at, to strengthen the combination of precision and not to improve the net-work or to increase the data sets. It is on the basis of this, we propose a judgement to distinguish data sets that different models are good at and an accurate segmentation method of MR brain tissue based on two existing deep learning models is proposed.\\

The remainder of this paper is organized as follows. In section 2, we present our method. Experiments and results are detailed in section 3. Finally, the discussion and the main conclusions are presented in section 4 and section 5 separately.

\section{Method}
In this paper, we propose our method based on the modified U-Net\cite{Deng2018Strategy} and VoxRes-Net\cite{Chen2017VoxResNet}. The process of the proposed method is below.
\begin{figure}[H]
	\begin{center}
		\includegraphics[width=1.\linewidth]{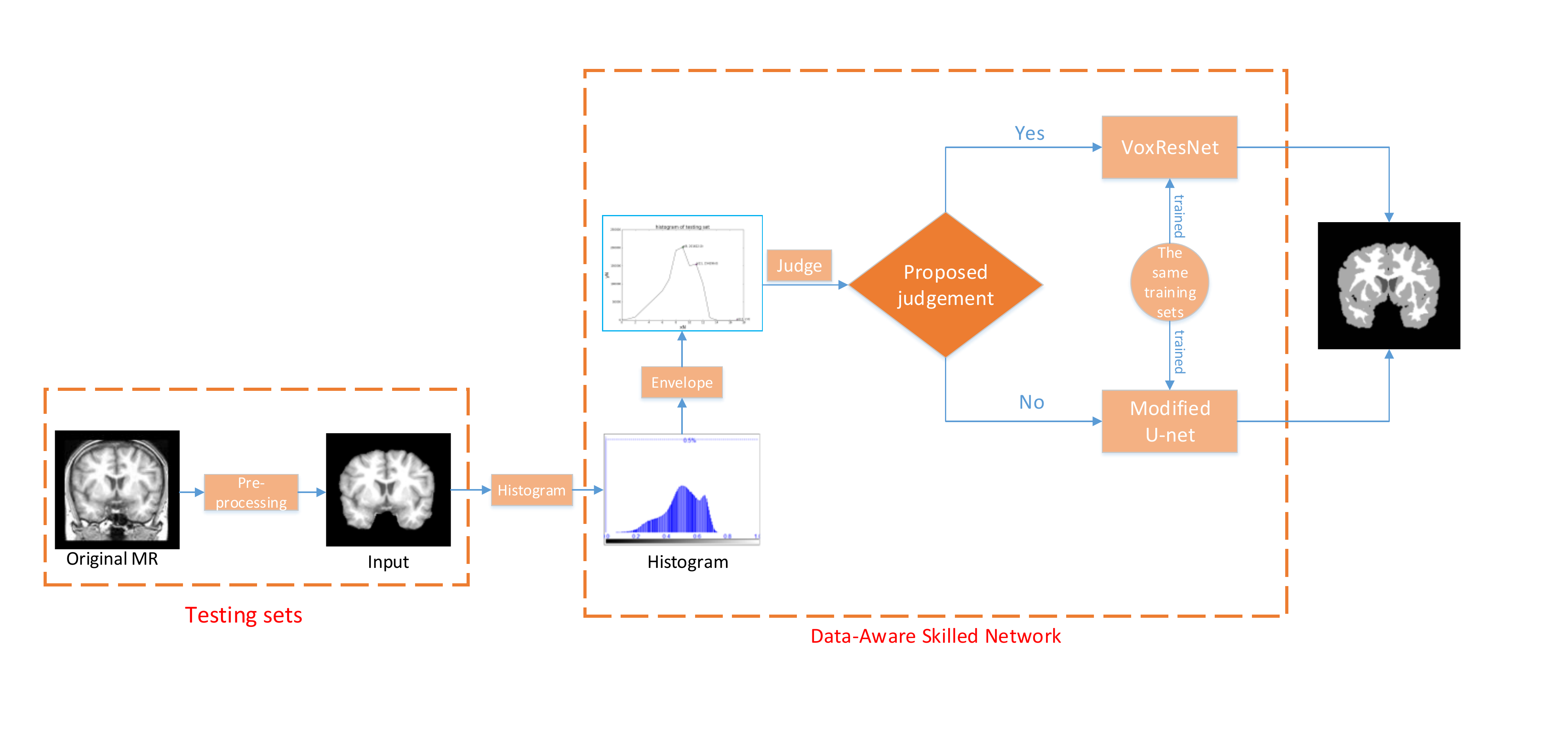}
		\caption{The overview of Data-Aware Skilled Network}
	\end{center}
\end{figure}
\subsection{Pre-processing} Firstly, the N4ITK is used to correct the bias field in each MRI sequence and the in-tensities are linearly transformed to [0, 1]. To limit the number of voxels considered in the classification, brains masks were generated with BET \cite{Smith2002Fast}. In each of the experi-ments, the samples from the training images were only selected from within the brain masks volumes. For each test image, only voxels within the brain mask volume were considered in the classification. \\

\subsection{Judgement}	
We calculate the histogram of each volume and judge whether this volume will be segmented better with the VoxResNet model\cite{Chen2017VoxResNet}. The judgement is below.

\begin{equation}
x_n-x_{n-1}> \frac{1}{x_N}
\end{equation}
\begin{equation}	
y_n\ge0.8*y_{n-1}
\end{equation}
\begin{equation}
x_n \le0.8*x_N 
\end{equation}\\

while ($x_n,y_n$) is the coordinate of peak point $P_n$,n is the amount of peak(always n $\ge2$),($0,x_N$) is the range of the abscissa.

The schematic diagram is as follows.
\begin{figure}[H]
\centering
	\includegraphics[width=1.\linewidth]{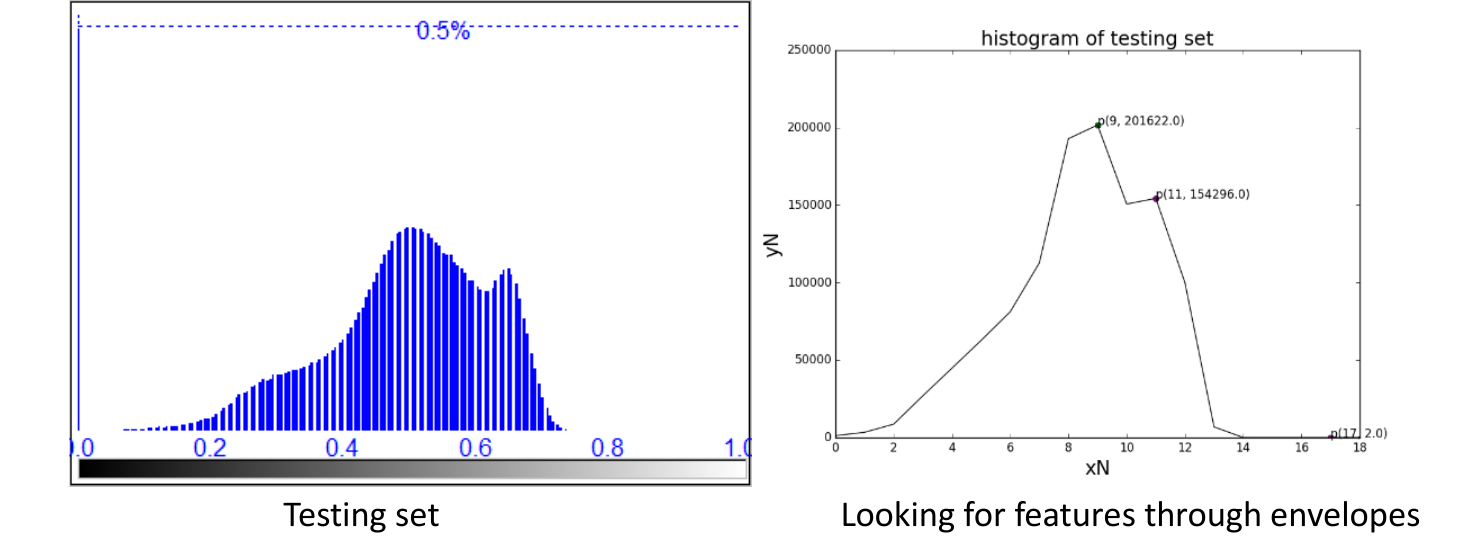}
\caption{The process to find the more valuable data from training dataset}
\end{figure}
If the input volume meets the judgements above, we sent it to the VoxResNet model and if not, we sent it to the modified U-Net model. It is important to note that the two models have been trained on the same training sets.
\subsection{Network architecture}
\subsubsection{Modified U-net\cite{Deng2018Strategy}}
\begin{figure}[H]
\begin{center}
\includegraphics[width=1.\linewidth]{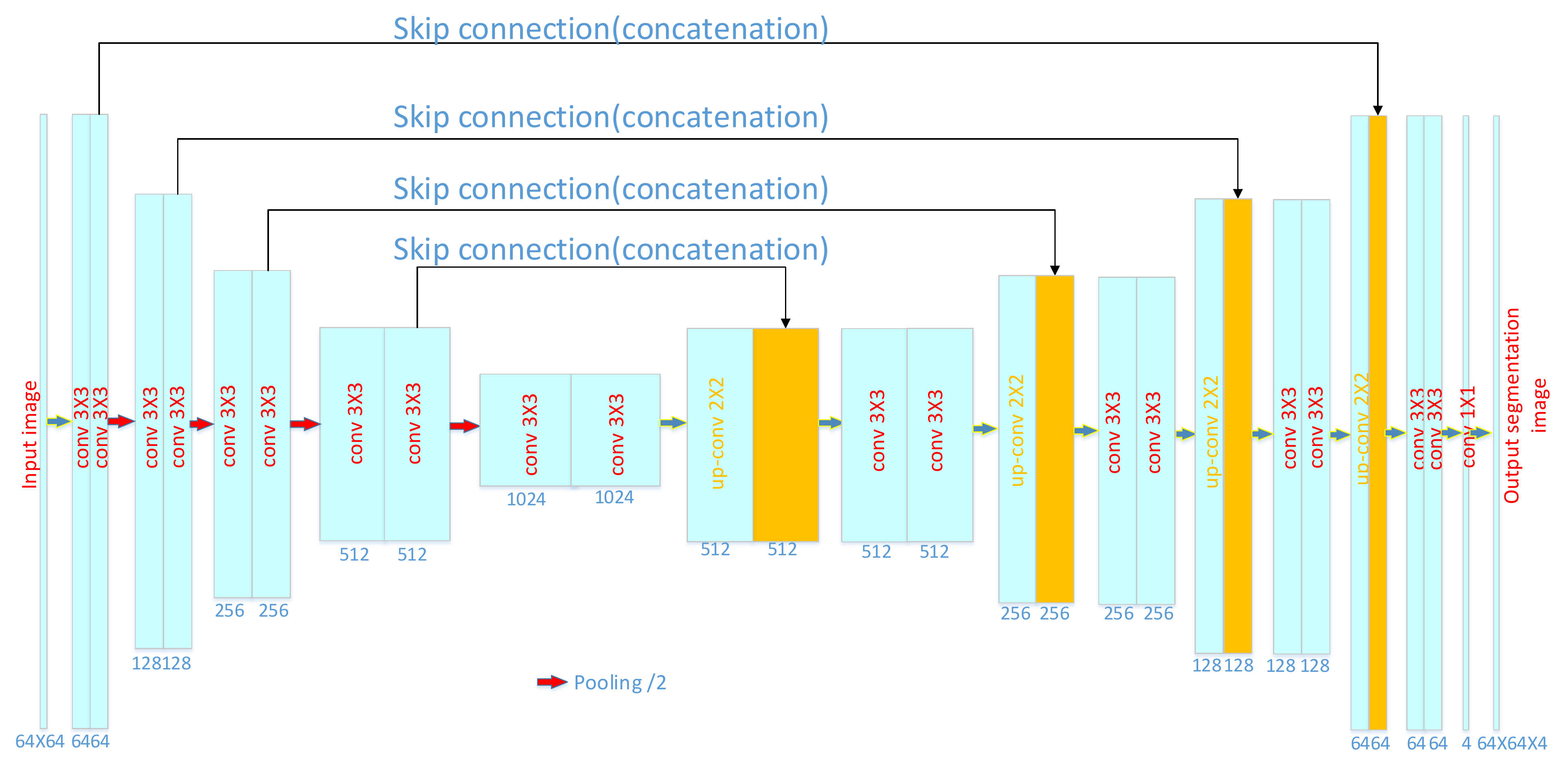}
\caption{The process to find the more valuable data from unlabeled dataset}
\end{center}
\end{figure}

Different from the purely U-net\cite{Ronneberger2015U}, our network can segment CSF, GM and WM three tissues at once because we use our own loss function below.
\begin{equation}
L(y,y')=4-\sum_{i=0}^{3}DSC(y_i,y_i')
\end{equation}

where $y_i$ and $y_i'$ are predicted and ground-truth for class i, respectively.

In the stage of segmentation reconstruction, we found the maximum probability among four classes and returned the corresponding label for each voxel rather than finding the optimized threshold.
At the same time, the shape of output is the same as input owing to the use of padding.

\subsubsection{VoxResNet \cite{Chen2017VoxResNet}}

\begin{figure}[H]
	\begin{center}
		\includegraphics[width=1.\linewidth]{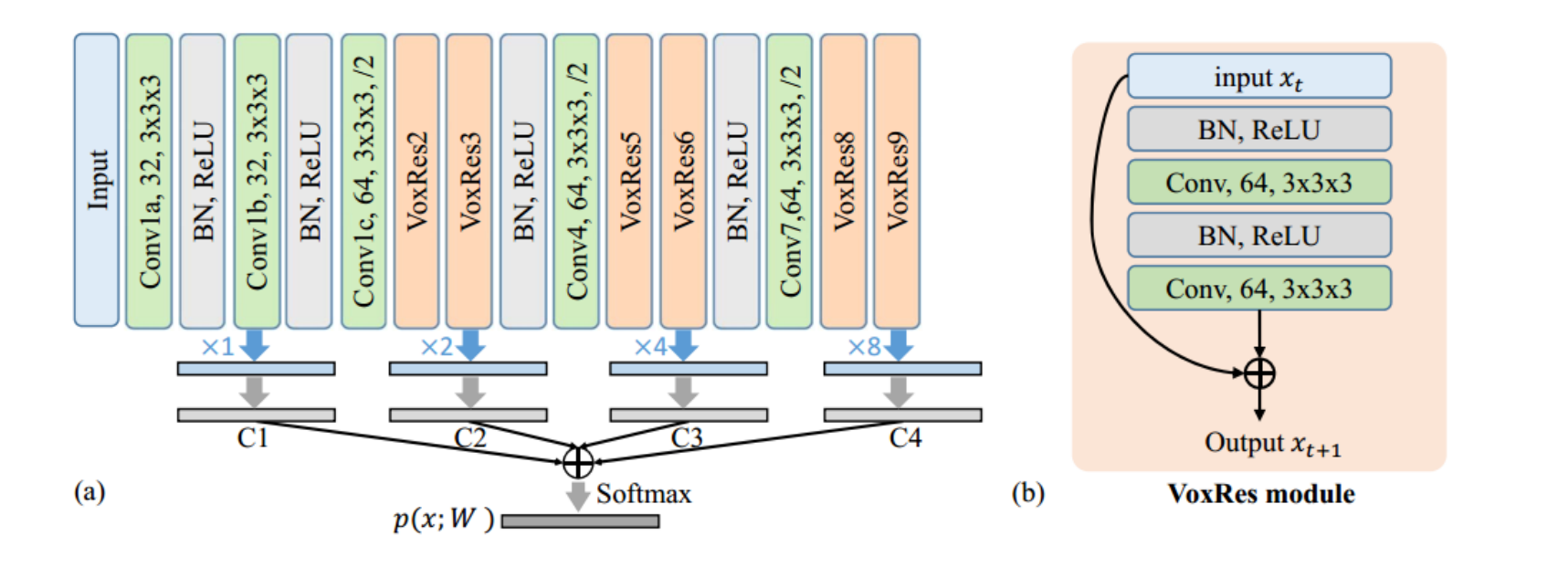}
		\caption{The process to find the more valuable data from unlabeled dataset}
	\end{center}
\end{figure}

It consists of stacked residual modules (i.e., VoxRes module) with a total of 25 volu-metric convolutional layers and 4 deconvolutional layers, which is the deepest 3D convolutional architecture so far. The details can be found in \cite{Chen2017VoxResNet}.

\section{Rusults}
We validate our method on the widely used IBSR 18 dataset\footnote{\url{https://www.nitrc.org/frs/?group_id=48}}. To avoid accidental error, we train the two model on several different training sets. The evaluation criteri-on is the three brain tissue (including CSF、GM、WM) average DCS (dice coeffi-cient), which is defined by:
\begin{equation}
DSC=\frac{2TP}{2TP+FP+FN} 
\end{equation}
Where TP, FP and FN are the subjects of true positive, false positive and false nega-tive predictions for the considered class.

The training set is the same for the two model and the testing subjects is the remain-ing except for the training set. For examples, when we use training set 1, the testing set is subject 6,7,8,9,10,11,12,13,14,15,16,17,18. (training set 1 including subject 1,2,3,4,5;training set 2 including subject 1,2,4,5,14;training set 3 including subject 10,11,12,13,14 and training set 4 including subject 1,2,3,4,5,10,11,12,13,14.). The metric we used here is the total average DSC of the three tissue of testing set tested by corresponding training sets.  

\begin{table}[H]
	\begin{tabular}{ccccc}
		\toprule
		Method& Training set 1& Training set 2&Training set 3&Training set 4\\
		\midrule
		VoxResNet\cite{Chen2017VoxResNet}& 0.8552& 0.8393&0.8439&0.8692\\
		Modified U-net\cite{Deng2018Strategy}& 0.8645& 0.8514&0.8771&0.8582\\
		Ours& \textbf{0.8713}& \textbf{0.8638}&\textbf{0.8796}&\textbf{0.8806}\\
		\bottomrule
	\end{tabular}
	\caption{Testing sets result of different methods with different training sets}
\end{table}

\section{Discusion}
According to an ancient Chinese saying, sometimes a foot may prove short while an inch may prove long. Every model should has its own advantages and disadvantages. Finding the advantages of every model and making best use of the ad-vantages to achieve megamerger are good ways of thinking, which have broad prospects.\\

In this paper, we achieved promising segmentation results based on exiting two mod-els only by designing a simple judgement¬ of histogram. In our future research, we will use more deep learning models and make a more accurate judgment by calculate the features of every sequence like \cite{Yuan2011Brain}\cite{Zou2009CAD}  rather than the whole patient volume. More-over, we can propose more stable judgement such as texture features, shape features and spatial relationships to distinguish different data sets, which are special for skilled model.\\

We believe that our thinking is also applicable in other various domains not limited to MR brain tissue segmentation and expect more researches to make a contribution in this area.

\section{Conclusions}
In this paper, we design a data-aware skilled network that can select the data sets that different deep learning models are good at and achieve more accuracy segmen-tation without changing the net or enhancing the data.\\

\bibliographystyle{IEEEtranS}
\bibliography{3D-DRL}

\begin{thebibliography}{10}
\providecommand{\url}[1]{#1}
\csname url@samestyle\endcsname
\providecommand{\newblock}{\relax}
\providecommand{\bibinfo}[2]{#2}
\providecommand{\BIBentrySTDinterwordspacing}{\spaceskip=0pt\relax}
\providecommand{\BIBentryALTinterwordstretchfactor}{4}
\providecommand{\BIBentryALTinterwordspacing}{\spaceskip=\fontdimen2\font plus
\BIBentryALTinterwordstretchfactor\fontdimen3\font minus
  \fontdimen4\font\relax}
\providecommand{\BIBforeignlanguage}[2]{{%
\expandafter\ifx\csname l@#1\endcsname\relax
\typeout{** WARNING: IEEEtranS.bst: No hyphenation pattern has been}%
\typeout{** loaded for the language `#1'. Using the pattern for}%
\typeout{** the default language instead.}%
\else
\language=\csname l@#1\endcsname
\fi
#2}}
\providecommand{\BIBdecl}{\relax}
\BIBdecl

\bibitem{Bao2015Multi}
S.~Bao and A.~C.~S. Chung, ``Multi-scale structured cnn with label consistency
  for brain {MR} image segmentation,'' pp. 1--5, 2015.

\bibitem{Chen2017VoxResNet}
H.~Chen, Q.~Dou, L.~Yu, J.~Qin, and P.~A. Heng, ``Voxresnet: Deep voxelwise
  residual networks for brain segmentation from 3d {MR} images,''
  \emph{Neuroimage}, vol. 170, 2017.

\bibitem{Despotovi2015MRI}
I.~Despotovi?, B.~Goossens, and W.~Philips, ``{MRI} segmentation of the human
  brain: challenges, methods, and applications.'' \emph{Computational and
  Mathematical Methods in Medicine,2015,(2015-3-1)}, vol. 2015, no.~6, pp.
  1--23, 2015.

\bibitem{Greenspan2006Constrained}
H.~Greenspan, A.~Ruf, and J.~Goldberger, ``Constrained gaussian mixture model
  framework for automatic segmentation of {MR} brain images,'' \emph{IEEE
  Transactions on Medical Imaging}, vol.~25, no.~9, pp. 1233--45, 2006.

\bibitem{Marroquin2002An}
J.~L. Marroquin, B.~C. Vemuri, S.~Botello, F.~Calderon, and A.~Fernandezbouzas,
  ``An accurate and efficient bayesian method for automatic segmentation of
  brain {MRI},'' \emph{IEEE Transactions on Medical Imaging}, vol.~21, no.~8,
  p. 934, 2002.

\bibitem{Moeskops2016Automatic}
P.~Moeskops, M.~A. Viergever, A.~M. Mendrik, L.~S. de~Vries, M.~J. Benders, and
  I.~Isgum, ``Automatic segmentation of {MR} brain images with a convolutional
  neural network.'' \emph{IEEE Transactions on Medical Imaging}, vol.~35,
  no.~5, pp. 1252--1261, 2016.

\bibitem{Nie2015FULLY}
D.~Nie, L.~Wang, Y.~Gao, and D.~Shen, ``Fully convolutional networks for
  multi-modality isointense infant brain image segmentation,'' \emph{Proc IEEE
  Int Symp Biomed Imaging}, vol. 108, pp. 1342--1345, 2015.

\bibitem{Qin2017Brain}
Z.~Qin, F.~Wang, Z.~Xiao, T.~Lan, and Y.~Ding, ``Brain tissue segmentation with
  the gka method in {MRI},'' in \emph{IEEE International Conference on Signal
  and Image Processing}, 2017, pp. 273--276.

\bibitem{Ronneberger2015U}
O.~Ronneberger, P.~Fischer, and T.~Brox, \emph{U-Net: Convolutional Networks
  for Biomedical Image Segmentation}.\hskip 1em plus 0.5em minus 0.4em\relax
  Springer International Publishing, 2015.

\bibitem{Smith2002Fast}
S.~M. Smith, ``Fast robust automated brain extraction,'' \emph{Human Brain
  Mapping}, vol.~17, no.~3, pp. 143--155, 2002.

\bibitem{Deng2018Strategy}
D.~Yang, S.~Yao, Y.~Zhu., M.~Zhu., and Y.~Kehong, ``A strategy of {MR} brain
  tissue images¡¯ suggestive annotation based on modified {U-Net},''
  \emph{arXiv:1807.07510[cs.CV]}, 2018.

\bibitem{Yuan2011Brain}
K.~Yuan, Z.~Tian, J.~Zou, Y.~Bai, and Q.~You, ``Brain {CT} image database
  building for computer-aided diagnosis using content-based image retrieval,''
  \emph{Information Processing \& Management}, vol.~47, no.~2, pp. 176--185,
  2011.

\bibitem{Zhang2015Deep}
W.~Zhang, R.~Li, H.~Deng, L.~Wang, W.~Lin, S.~Ji, and D.~Shen, ``Deep
  convolutional neural networks for multi-modality isointense infant brain
  image segmentation,'' \emph{Proc IEEE Int Symp Biomed Imaging}, vol. 108, pp.
  1342--1345, 2015.

\bibitem{Zou2009CAD}
J.~Zou, ``Reseach on content-based brain {CT} image retrieval for computer
  aided diagnosis,'' \emph{Master's degree thesis of Tsinghua University},
  2009.

\end{thebibliography}

\end{document}